\documentclass[conference]{IEEEtran}
\IEEEoverridecommandlockouts
\usepackage{cite}
\usepackage{amsmath,amssymb,amsfonts}
\usepackage{algorithmic}
\usepackage[dvipdfmx]{graphicx}
\usepackage[dvipdfmx]{color}
\usepackage{subcaption}
\usepackage{textcomp}
\usepackage{bm}
\usepackage{xcolor}
\usepackage{balance}
\def\BibTeX{{\rm B\kern-.05em{\sc i\kern-.025em b}\kern-.08em
    T\kern-.1667em\lower.7ex\hbox{E}\kern-.125emX}}
\makeatletter
\newcommand{\newlineauthors}{%
  \end{@IEEEauthorhalign}\hfill\mbox{}\par
  \mbox{}\hfill\begin{@IEEEauthorhalign}
}
\makeatother
\begin{document}

\title{
Robotic Stroke Motion Following the Shape of \\the Human Back: Motion Generation and \\Psychological Effects
{
}
\thanks{This work was supported by JST CREST Number JPMJCR17A5, Japan, and conducted mainly at the Division of Information Science, Nara Institute of Science and Technology, Japan.}
}

\author{\IEEEauthorblockN{Akishige Yuguchi}
\IEEEauthorblockA{\textit{Faculty of Advanced Engineering} \\
\textit{Tokyo University of Science}\\
Katsushika-ku, Tokyo, Japan \\
akishige.yuguchi@rs.tus.ac.jp}
\and
\IEEEauthorblockN{Tomoki Ishikura}
\IEEEauthorblockA{\textit{Division of Information Science} \\
\textit{Nara Institute of Science and Technology}\\
Ikoma, Nara, Japan}
\and
\IEEEauthorblockN{Sung-Gwi Cho}
\IEEEauthorblockA{\textit{School of Science and Engineering} \\
\textit{Tokyo Denki University}\\
Hiki-gun, Saitama, Japan \\
s.g.cho@mail.dendai.ac.jp}
\newlineauthors
\IEEEauthorblockN{Jun Takamatsu}
\IEEEauthorblockA{\textit{Applied Robotics, Microsoft} \\
Redmond, WA, USA \\
jun.takamatsu@microsoft.com}
\and
\IEEEauthorblockN{Tsukasa Ogasawara}
\IEEEauthorblockA{
\textit{Nara Institute of Science and Technology}\\
Ikoma, Nara, Japan \\
ogasawar@is.naist.jp}
}
\maketitle


\section{Introduction}
Several caregiving techniques such as \textit{Tactile Massage}~\cite{tactilemassage} and \textit{Humanitude}~\cite{humanitude} require skillful or therapeutic touch through physical interaction.
Furthermore, several studies have reported that touching and stroking patients with such skills \textit{e.g.}, massage produces several kinds of positive effects such as a reduction in pain~\cite{Goldstein2016} and improvements in sleep quality~\cite{andersson2009tactile}.
On the other hand, the lack of caregivers and the labor shortage in those skills have been increasing, hence, it is expected that robot arms instead of human hands perform massage\footnote{Aescape Inc., https://www.aescape.com/} and dementia care through skillful stroke touch~\cite{sumioka2021}. 

To address those expectations, a few studies elucidated that stroking the human back using a robot arm evokes pleasant feelings in participants similar to those evoked by humans~\cite{ishikura2023,ishikura2024}. 
While the robotic stroke motions on the back were implemented with a linear trajectory for vertically stroking the back in these studies, the motion appearance was far from that of the stroke motions by humans because the human hand can follow the shape of the human back by stroking. 
We thought that it remains unclear whether the robot arm can generate stroke motions following the shape of the human back as well as humans do and whether the robotic stroke motions following the shape of the back evoke pleasant feelings in participants similar to those evoked by humans.

In this study, we propose a method to generate stroke motions using a robot arm to follow the shape of the human back. 
The method generates a trajectory to follow the shape of the human back for a robotic stroke motion with a cubic function to model the shape of the human back from the depth image obtained by a depth camera.
We confirm whether the generated trajectories can be similar to the trajectories of the stroke motions on the back by humans.
Finally, we evaluate the psychological effect on the participants' backs by comparing the robotic stroke motion following the shape of the human back with the proposed method and the conventional stroke motion with a linear trajectory.


\section{Trajectory Generation for Robotic Stroke Motion Following the Shape of the Human Back}
\label{sec:method}
The stroke motion in this study refers to a stroke motion in a fixed vertical direction from the upper human back to the lower human back, as in the aforementioned previous studies~\cite{ishikura2023,ishikura2024}.
To generate a trajectory of the stroke motion following the shape of the human back, we use a depth camera with calibrated internal parameters to recognize the shape of the back in the stroking direction by calculating the 3D position in actual space from a depth image.
Specifically, first, we select the start and end positions of the stroke motion from the obtained depth image, connect the start and end positions of the stroke motion by a straight line, and then store the obtained depth values as $z$ in the section between the start and end positions from 3D positions~$p(x_i, y_i, z_i)$ in the camera coordinate system~$\bm{\Sigma}_C$. 
$x$ is the horizontal direction of the captured depth image (the width) and $y$ is the vertical direction of the captured image (the height). 
In this case, since the stroking direction is constrained only in the vertical direction (on the~$y$ axis), we set the $x$ values at the start and end positions to be the same in the image.

Next, we generate a trajectory for the stroke motion using a robot arm from the recognized shape of the human back with the following four steps: 

\begin{enumerate}
 \item Convert the recognized shape of the human back to approximate a cubic curve
 \item Calculate the waypoint positions 
 for the trajectory of the stroke motion from the approximated curve
 \item Calculate the target angle for the orientation control of the end-effector
 \item Transform the waypoint positions 
 from the camera coordinate system~$\bm{\Sigma}_C$ to the robot coordinate system~$\bm{\Sigma}_R$
\end{enumerate}
First, we convert the recognized shape of the back by approximating a cubic function~($z=ay^3+by^2+cy+d$) into a smooth curve.
Second, we calculate the corresponding depths of the height~$y$ in increments of 1.0 [mm] from the approximate curve and then set those points as the waypoints~$p_\text{way}(x, y_i, z_i)$ through which the end-effector moves.
Third, we calculate the target angle~$\theta_{\text{target}_x}$ by~\eqref{angle} to control the orientation of the end-effector in the~$x$-axis direction.
\begin{equation}
\label{angle}
  \theta_{\text{target}_x} = \tan^{-1}\frac{z_i - z_{i-1}}{y_i - y_{i-1}}
\end{equation}
Finally, we transform the waypoint positions $p_\text{way}(x,y_i,z_i)$ from the camera coordinate system~$\bm{\Sigma}_C$ to the robot coordinate system~$\bm{\Sigma}_R$.
We input the generated waypoints~$p_\text{way}(x, y_i,z_i)$ and the target orientation~$\theta_{\text{target}_x}$ of the end-effector into the motion planning tool Moveit\footnote{MoveIt, https://moveit.ros.org} to generate the stroke motion.

\section{Generation of Robotic Stroke Motion and Its Trajectory Evaluation}\label{sec:robot}
\subsection{Trajectory and Motion Generation}
We used a cooperative robot arm UR3e produced by Universal Robots similar to those used in the previous studies~\cite{ishikura2023,ishikura2024}.
We attached a depth camera Intel RealSense D435 near the robot base.
To generate the stroke motions on the back, we used a mannequin dressed in patient wear\footnote{SG1441, Nagai Leben, Inc.}.

\begin{figure}[t]
  \begin{minipage}{0.49\hsize}
  \begin{center}
   \includegraphics[width=\linewidth]{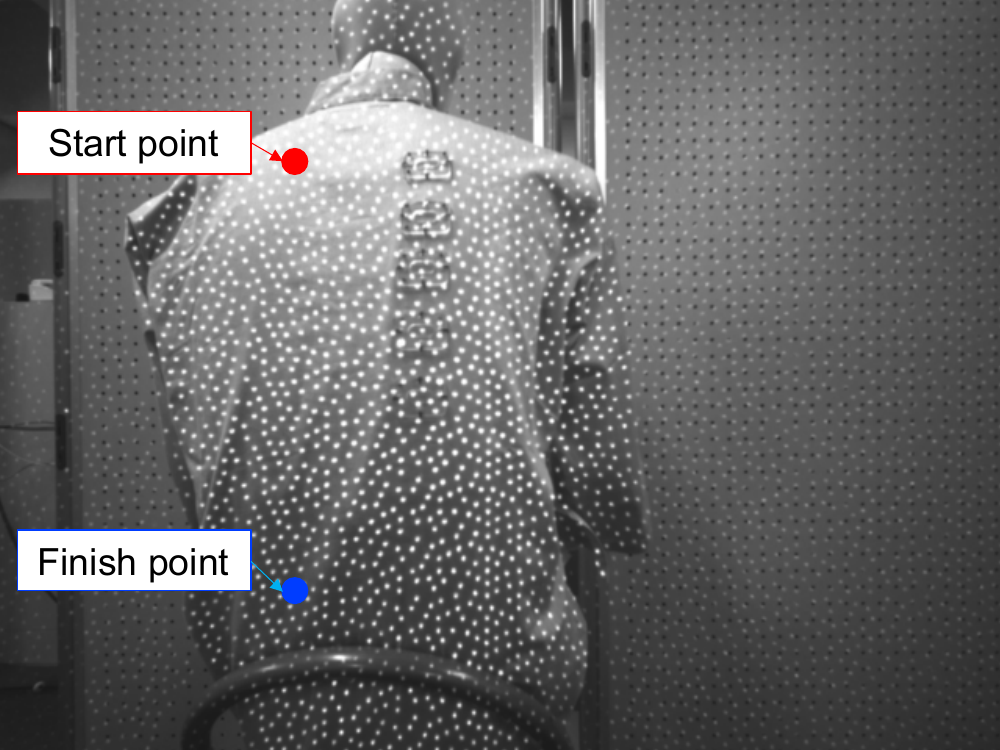}\\
   A. Camera image. 
  \end{center}
 \end{minipage}
 \begin{minipage}{0.49\hsize}
  \begin{center}
   \includegraphics[width=\linewidth]{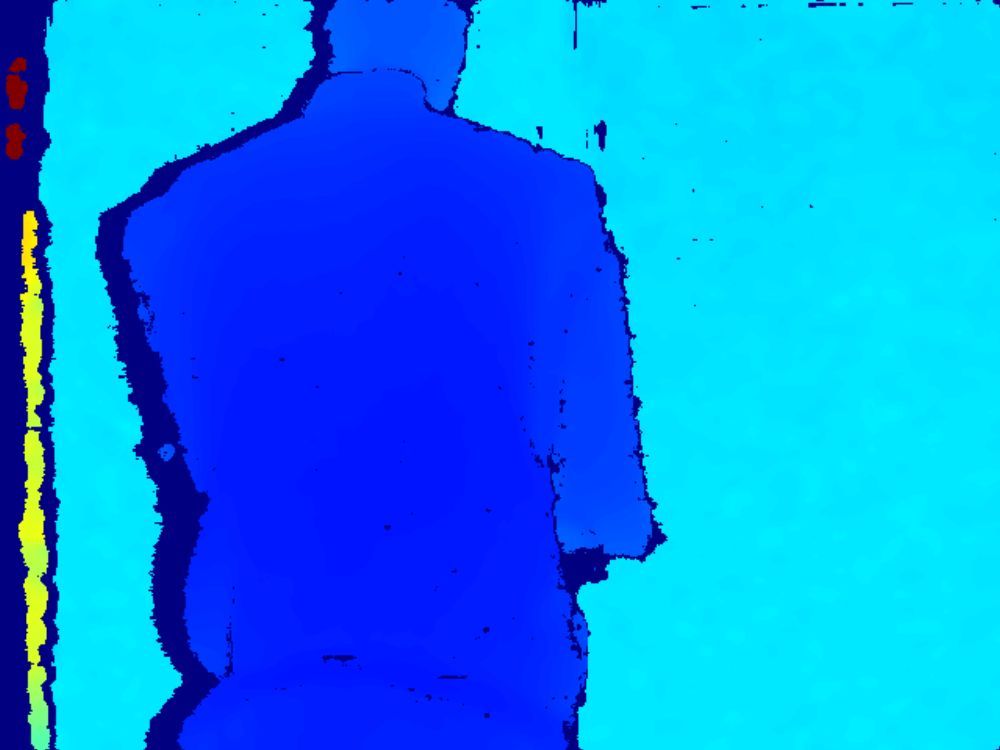}\\
   B. Depth image. 
  \end{center}
 \end{minipage}
 \caption{Actual camera and depth images of the back.}
 \label{depth}
\end{figure}

In the actual shape recognition of the human back, we took a depth image of the back and then selected the start and end positions of the stroke motion. 
Then, we generated the trajectory and the stroke motion using the proposed method.
Figure~\ref{depth} shows the actual recognized shape of the back and the start and end positions.
\begin{figure*}[t]
  \centering
          \includegraphics[width=0.85\linewidth]{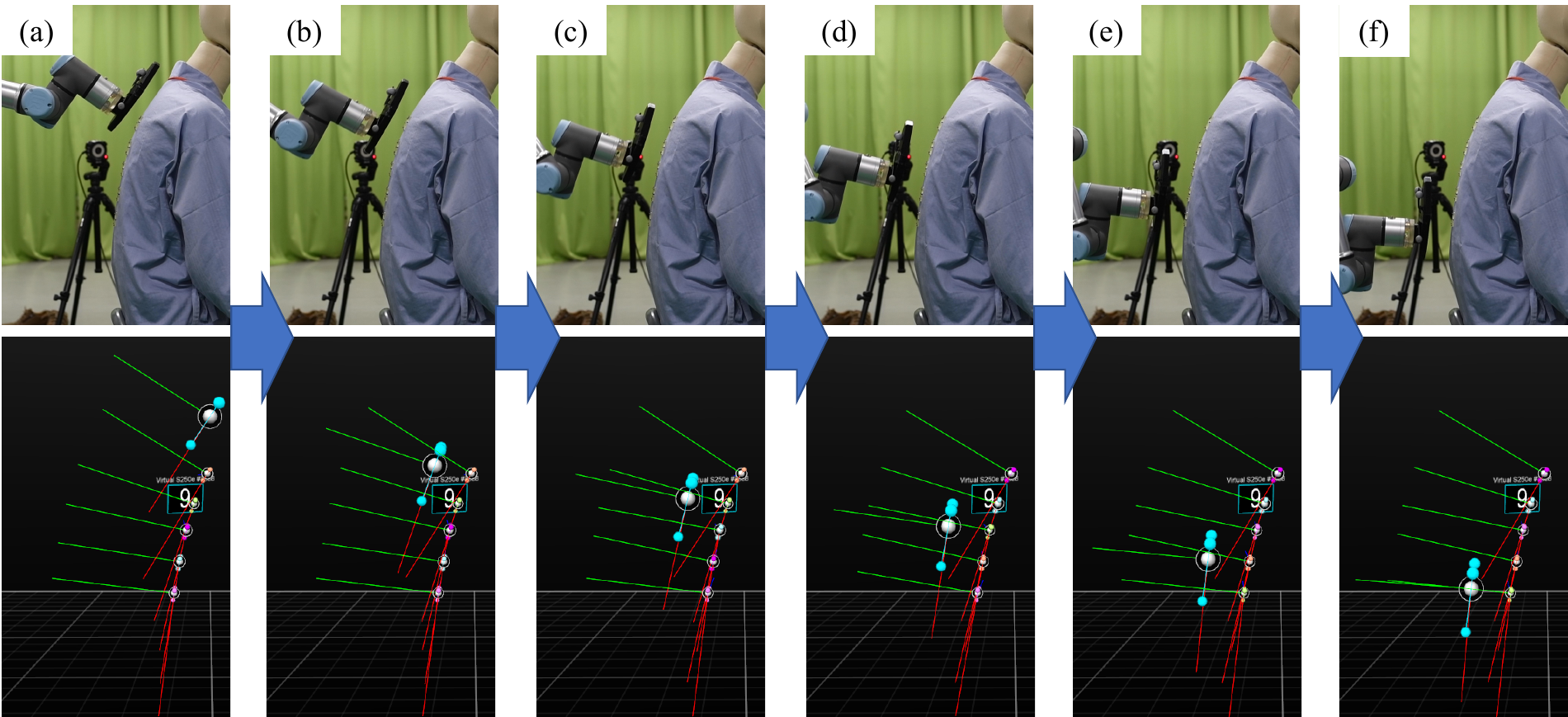}
        \caption{The actual robotic stroke motion and the measured normals on the back and robot's end effector.}
        \label{robot_normal}
\end{figure*} 
\begin{figure*}[t]
  \centering
        \includegraphics[width=0.85\linewidth]{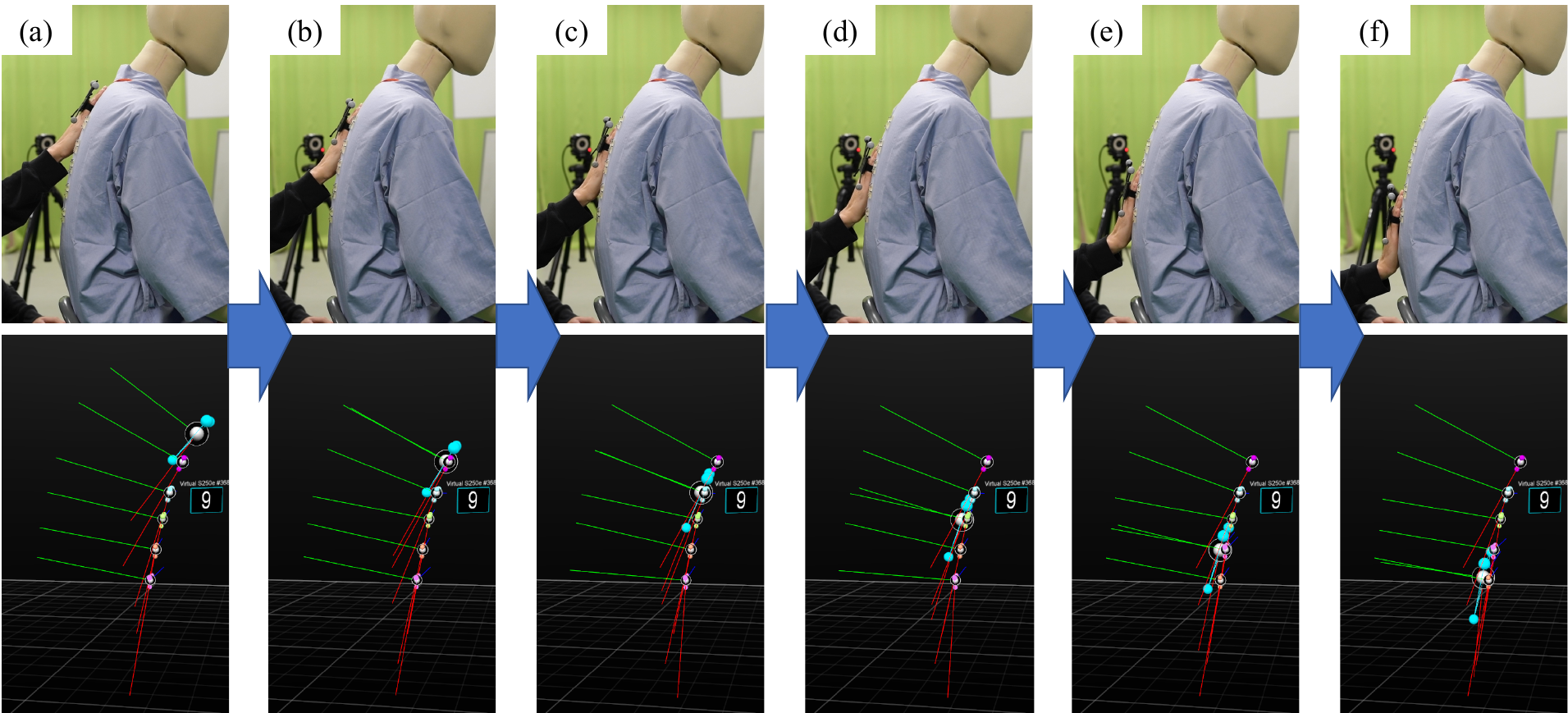}
        \caption{The actual human stroke motion and the measured normals on the back and hand.}
        \label{human_normal}
\end{figure*} 

\subsection{Evaluation}
To evaluate the generated trajectory of the robotic stroke motion following the shape of the human back, 
we calculated how much the normal directions match from the normal vector~$\bm{B}$ of the back and the normal vector~$\bm{E}$ of the end-effector by~\eqref{norm}.
\begin{equation}
\label{norm}
  \theta_{\text{error}_x} = \cos^{-1}\frac{\bm{B}\cdot\bm{E}}{|\bm{B}||\bm{E}|}.
\end{equation}
We measured the generated motion trajectory five times using an optical motion capture system OptiTrack\footnote{OptiTrack, http://www.optitrack.com}.
The normal vector $\bm{B}$ was the vertical direction on the plane of the rigid body created from the markers on the back.
We attached five pairs of the markers at 5.0 [cm] increments from the upper back of the patient wear worn by the mannequin, 20.0 [cm] in total.
The normal vector~$\bm{E}$ was the vertical direction on the plane of the rigid body created from the markers on the end-effector.
The trajectory range of the stroke motion was defined from the uppermost markers on the upper back to passing through the lowest markers.

Figure~\ref{robot_normal} shows the actual generated motion using the robot arm, the back displayed on the motion capture system, and the transition of the normal vector~$\bm{E}$ of the end-effector.
The average error~$\bar{\theta}_{\text{error}_x}$ at each point of the motion trajectories was 5.97 [deg] and the maximum error~$\max{\theta_{\text{error}_x}}$ was 8.26 [deg].

\section{Comparison with Stroke Motion Trajectory to the Back by Humans}\label{sec:human}
\subsection{Measurement of Stroke Motions by Humans}
We measured stroke motions on the back by humans to compare the trajectories of the robotic stroke motions following the back with those of the motions by humans.

For the stroke motions by humans, we measured the demonstration of the motion by two performers who stroked the participants' backs in the previous studies~\cite{ishikura2023,ishikura2024}.
We measured the vertical direction on the plane of the rigid body created from the markers attached to the tip of the performer's right hand as the normal vector~$\bm{H}$ of the hand.
Then, we evaluated how similar the vector~$\bm{H}$ replaced from the normal vector~$\bm{E}$ in~\eqref{norm} is to the normal vector~$\bm{B}$ of the back.
In addition to the measurement in Section~\ref{sec:robot}, we measured the motions on the back of a mannequin wearing the patient wear with the markers in the same posture five times, starting from the uppermost markers on the upper back to passing through the lowest markers.



\subsection{Measurement Result}
Figure~\ref{human_normal} shows the actual human demonstration of the stroke motion, the back displayed on the motion capture system, and the transition of the normal vector~$\bm{H}$ of the hand.

The average error~$\bar{\theta}_{\text{error}_x}$ from each point on all the motion trajectories was 4.78 and 5.52 [deg] for the first and second performers, respectively, and the maximum error~$\max{\theta_{\text{error}_x}}$ was 7.30 and 8.13 [deg] for the first and second performers, respectively.
These results indicated that the average error~$\bar{\theta}_{\text{error}_x}$ and the maximum error~$\max{\theta_{\text{error}_x}}$ of the robotic stroke motion in Section~\ref{sec:robot} were 5.97 [deg] and 8.26 [deg], which were close to those of the stroke motions by humans. 
Therefore, we deemed that the robotic stroke motion using the proposed method could follow the shape of the back. 



\section{Subjective Experiment}
\subsection{Method}
To evaluate the effect of the stroke motion following the shape of the human back using the actual robot arm, we attached the human-mimetic robotic hand used in the previous studies~\cite{ishikura2023,ishikura2024} to the end-effector, and compared the psychological effect of the stroke motion with the proposed method and the conventional stroke motion with a linear trajectory used in the previous studies by subjective evaluation.

From the previous studies, we used the Affect Grid~\cite{grid} for subjective evaluation.
Affect Grid consists of 9 $\times$ 9 two-dimensional grids with a horizontal and vertical axis, each representing a 9-point scale of valence (pleasant-unpleasant) and arousal level (high-low). 
The left and right directions in the horizontal axis correspond to pleasant and unpleasant feelings (valence), and the up and down directions in the vertical axis correspond to high and low arousal.

We prepared totally four conditions: two types of motions,~\textit{i.e.,} the stroke motion following the shape of the human back (the proposed method) and the stroke motion with a linear trajectory (the conventional method used in the previous studies), and two types of speeds (slow: 2.8~[cm/s] and medium: 8.5~[cm/s]) similar to the experiments in the previous studies.

To control the experimental environment, we asked participants to wear patient wear and earmuffs as in the previous studies. 
Furthermore, to reduce the participants' anxiety, we asked the participants to observe only the approach to the back before the start of the robotic stroke motion through a screen placed in front of the chair for the participants.

In the experimental procedure, first, we explained the experimental procedure to a participant 
and then asked him to wear the patient wear.
Second, we took an image of the shape of the participant's back with the depth camera and asked him to move as little as possible during the experiment.
The shape of the back was recognized only once before the start of the experiment, and the stroke motion following the shape as the condition was generated from this shape recognition result.
Third, we conducted a pretest that the robot arm generated the stroke motion following the participant's back at low speed to reduce the participant's anxiety about the robot arm and to confirm the visibility of the approach.
After the pretest, we repeated the following four steps 20 times.







\begin{figure}[t]
  \begin{center}
   \includegraphics[width=\linewidth]{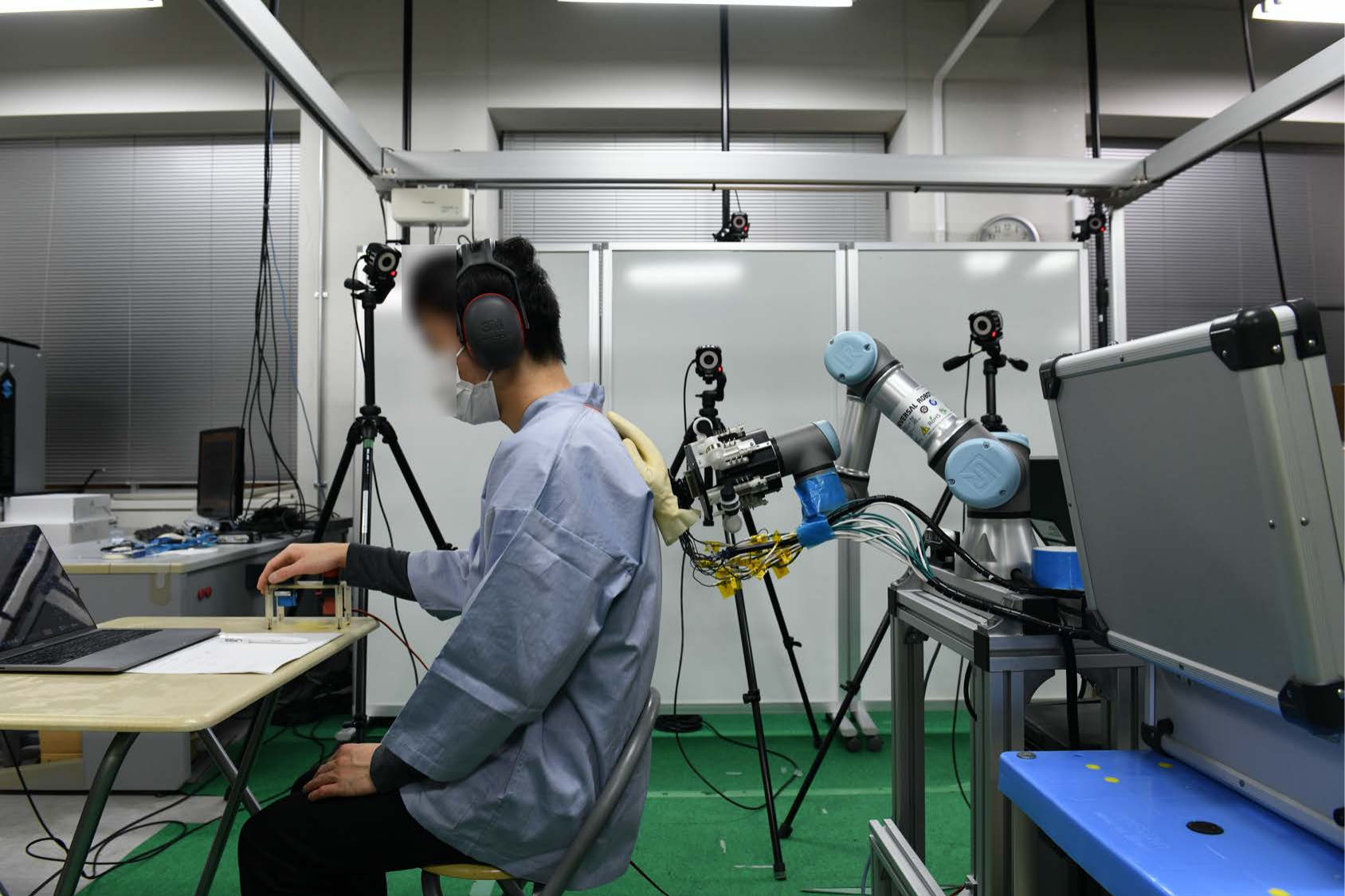}\\
  \end{center}
 \caption{The actual experimental setup}
 \label{motion_experimentroom}
\end{figure}

\begin{enumerate}
    \item We randomly select one of the four conditions, combining the speed (low and medium speed) and the motions (the stroke motion following the shape of the human back and the stroke motion with a linear trajectory).
    \item The robot arm approaches the participant's back while showing the approach on the participant's front screen.
    \item We hide the video on the front screen and the stroke motion is performed for about 20 seconds.
    \item After the stroke motion, the participant intuitively and subjectively evaluates the feeling by the stroke motion in about 10 seconds.
\end{enumerate}

\begin{figure}[t]
  \begin{center}
    \begin{tabular}{c}
      \begin{minipage}{\hsize}
          \centering
          \includegraphics[width=1\linewidth]{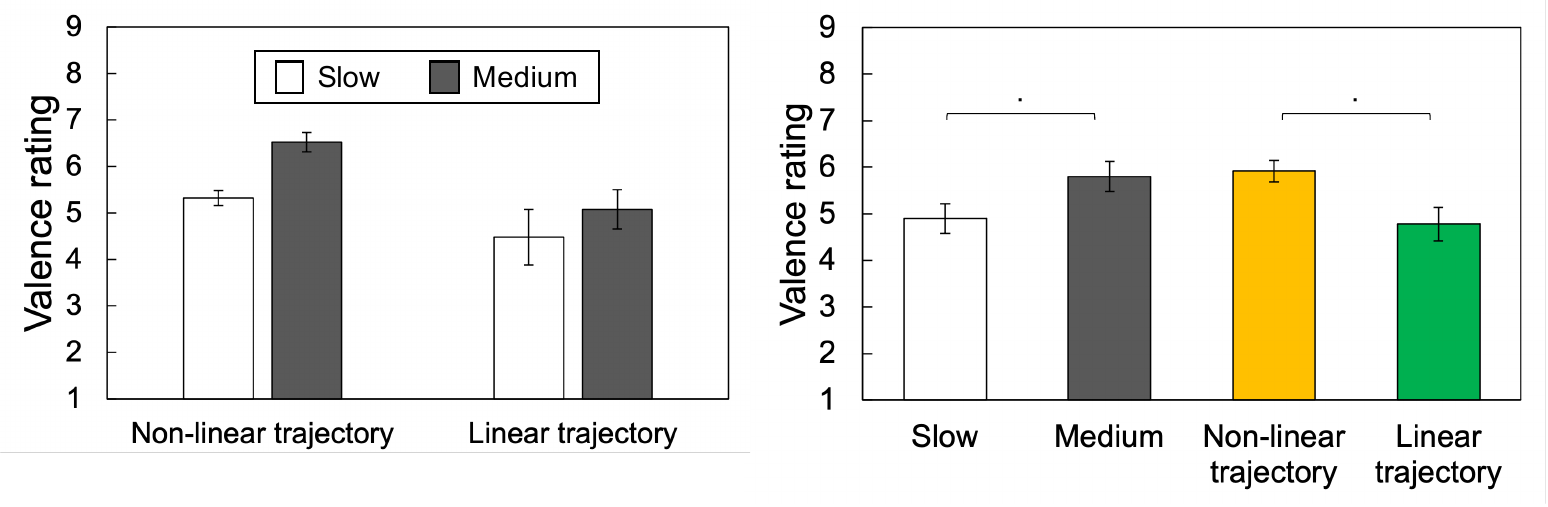}
      \end{minipage}\\
      A. Valence.\\
      \begin{minipage}{0.65\hsize}
          \centering
          \includegraphics[width=0.75\linewidth]{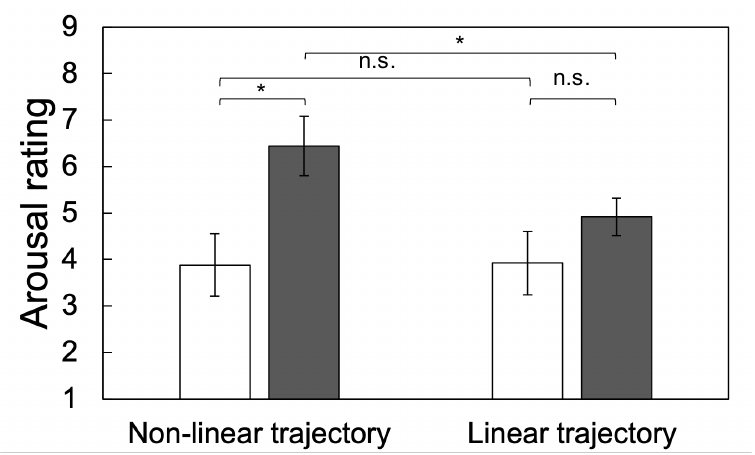}
      \end{minipage}\\
      B. Arousal.\\
    \end{tabular}
    \caption{The result of subjective evaluation using Affect Grid ($*$: p $< 0.05$). Note that the condition~\textit{non-linear trajectory} is the stroke motion following the shape of the human back with the proposed method.}
    \label{subjective_mosion}
  \end{center}
\end{figure}
\subsection{Result}
This experiment was approved by the Ethics Committee of Nara Institute of Science and Technology, Japan.
Five healthy Japanese adult males (mean~$\pm$~SD age, $23.4~\pm~0.8$ years) participated in the experiment after informed consent was obtained from all participants.

Figure~\ref{subjective_mosion} shows the results of the Affect Grid.
Figure~\ref{subjective_mosion}-A shows the valence scores (pleasant - unpleasant), with higher scores indicating pleasantness and lower scores indicating unpleasantness.
The left and right figures in Figure~\ref{subjective_mosion}-A show the valence scores for each condition and the two factors (speed and motion) for each level (speed: low and medium speed, motion type), respectively.
Figure~\ref{subjective_mosion}-B shows the arousal scores (high - low), with higher scores indicating higher arousal and lower scores indicating lower arousal.


In the valence scores, Figure~\ref{subjective_mosion}-A shows that the stroke motion following the shape of the human back evoked more pleasantness than the conventional stroke motion with a linear trajectory, and the difference in speed was similar to that in the previous studies.
An ANOVA analysis was conducted to evaluate the difference between the two motion types, and no interaction was found: $F(1,4) = 0.652, p > 0.1$.
Furthermore, a main effect test showed marginal significance with $F(1,4) = 7.236, p < 0.1 $ and $F(1,4) = 5.548, p < 0.1 $ for the types of motion and speed, respectively.

In the arousal scores, Figure~\ref{subjective_mosion}-B shows that the stroke motion following the shape of the human back evoked higher arousal than the conventional stroke motion with a linear trajectory, and the difference in speed was similar to that in the previous studies.
An ANOVA analysis revealed an interaction effect of $F(1,4) = 15.844, p < 0.5$.
Furthermore, a simple main effect test was conducted as a post-hoc test, and no significant difference was found in comparing the low-speed motions, $F(1,4) = 0.004, p > 0.1$. The comparison of the medium-speed motions was significantly different with $F(1,4) = 6.382, p < 0.05 $.

The speed comparison of the stroke motions following the shape of the human back showed a significant difference, $F(1,4) = 7.851, p < 0.05 $, indicating that the medium-speed stroke motion following the shape of the back evoked higher arousal than the low-speed stroke motion following the shape of the back.
On the other hand, no significant difference was found in the speed comparison of the stroke motions with a linear trajectory, $F(1,4) = 1.198, p > 0.1$.


\subsection{Discussion}
In terms of valence, the stroke motion following the shape of the human back tended to evoke more pleasantness than the conventional stroke motion with a linear trajectory.
On the speed aspect, it was shown that the medium-speed stroke motion tended to evoke more pleasant than the low-speed stroke motion, similar to the results in the previous studies.

In terms of arousal, the stroke motion following the shape of the human back tended to evoke higher arousal than conventional stroke motion with a linear trajectory.
In particular, the medium-speed stroke motion following the shape of the back was higher arousing than the medium-speed stroke motion with a linear trajectory.
On the speed aspect, it was shown that the medium-speed stroke motion following the shape of the back evoked higher arousal than the low-speed stroke motion following the shape of the back, similar to the previous studies.

These results indicated that the stroke motion following the shape of the human back with the proposed method tended to evoke pleasant and higher arousal, \textit{i.e.},~more pleasant and active feelings, than the conventional stroke motion with a linear trajectory, and that the medium-speed stroke motion especially evoked higher arousal, \textit{i.e.},~more active feelings.
The differences in the speed of the stroke motions were similar to those in the previous studies, confirming the previous findings.

The results suggested that the stroke motion following the shape of the human back with the proposed method tended to have a certain positive effect compared to the conventional stroke motion with a linear trajectory.
Thus, we expect that these human-like therapeutic stroke motions on the back could be implemented by robots instead of human skills to perform massage and dementia care.

On the other hand, note that this study has a few possible limitations.
First, we only used a cubic function to generate a trajectory of the robotic stroke motions following the shape of the human back. The actual shape of the human back is always not fixed \textit{e.g.},~when a human breathes. Hence, real-time generation of trajectories of the robotic stroke motions is needed.
We consider that there is a case where the only use of a cubic function may be about to fail the target trajectory generation.
Second, as referred to in Section~\ref{sec:method}, the stroke direction on the human back is constrained.
In actual situations of care and massage, we consider that the stroke direction can be various.
Third, the number of participants in the subjective experiment was very limited. To confirm the general psychological effects, we need to ask more participants to have experiences with the stroke motion with the proposed method.




\section{Conclusion}
In this study, to perform the robotic stroke motions following the shape of the human back similar to the stroke motions by humans, in contrast to the conventional robotic stroke motion with a linear trajectory, we proposed a trajectory generation method for a robotic stroke motion following the shape of the human back.
We confirmed that the accuracy of the method's trajectory was close to that of the actual stroking motion by a human.
Furthermore, we conducted a subjective experiment to evaluate the psychological effects of the proposed stroke motion in contrast to those of the conventional stroke motion with a linear trajectory.
The experimental results showed that the actual stroke motion following the shape of the human back tended to evoke more pleasant and active feelings than the conventional stroke motion.

In future work, we would like to increase the number of participants and further evaluate the psychological effects of the robotic stroke motion following the shape of the human back. 
In addition, we would like to extend our method to generate robotic stroke motions following the shapes of other parts of the body.


\bibliographystyle{IEEEtran}
\bibliography{main.bbl}
\end{document}